\documentclass[letterpaper, 10 pt, conference]{ieeeconf}  

\IEEEoverridecommandlockouts                              

\usepackage[utf8]{inputenc} 
\usepackage[T1]{fontenc}    
\usepackage{hyperref}       
\usepackage{url}            
\usepackage{booktabs}       
\usepackage{amsfonts}       
\usepackage{nicefrac}       
\usepackage{microtype}      
\usepackage{color}
\usepackage{float}

\usepackage{tabularx}

\usepackage{graphicx}
\usepackage{subcaption}

\title{LabelFusion: A Pipeline for Generating Ground Truth Labels for Real RGBD Data of Cluttered Scenes}

\author{
  Pat Marion*, Peter R. Florence*, Lucas Manuelli* and Russ Tedrake\\
  Computer Science and Artificial Intelligence Laboratory\\
  Massachusetts Institute of Technology\\
  \texttt{\{pmarion,peteflo,manuelli,russt\}@csail.mit.edu} \\
  $^*$\it{These authors contributed equally to this work.}\\
}

\begin{document}
\maketitle


\begin{abstract}


Deep neural network (DNN) architectures have been shown to
outperform traditional pipelines for object segmentation and pose estimation
using RGBD data, but the performance of these DNN pipelines is directly tied to how representative the training data is of the true data. Hence a key requirement for employing these methods
in practice is to have a large set of labeled data for your specific robotic manipulation task, a
requirement that is not generally satisfied by existing datasets. In this 
paper we develop a pipeline to rapidly generate high quality RGBD data with pixelwise labels and object poses. We use an RGBD camera
to collect video of a scene from multiple viewpoints and leverage existing reconstruction techniques to produce a 3D dense reconstruction.
We label the 3D reconstruction using a human assisted ICP-fitting of object meshes. By reprojecting the results of labeling the 3D scene
we can produce labels for each RGBD image of the scene. This pipeline enabled us to collect over 1,000,000 labeled object instances in just a few days. We use
this dataset to answer questions related to how much training data is required, and of what quality
the data must be, to achieve high performance from a DNN architecture. Our dataset and annotation pipeline are available at \texttt{labelfusion.csail.mit.edu}.

\end{abstract}



\section{Introduction}

Advances in neural network architectures for deep learning have made significant impacts on perception for
robotic manipulation tasks. State of the art networks are able to produce high quality pixelwise segmentations of RGB images, which can be used as a key component for 6DOF object pose estimation in cluttered environments \cite{Wong17,zeng2016multi}.  
However for a network to be useful in practice it must be fine tuned on
labeled scenes of the specific objects targeted by the manipulation task, and 
these networks can require tens to hundreds of thousands of labeled training examples
to achieve adequate performance. To acquire sufficient data for each specific robotics
application using once-per-image human labeling would be prohibitive, either in time or money.
While some work has investigated closing the gap with
simulated data \cite{richter2016playing,Johnson-Roberson:2017aa,RosCVPR16,hattori2015learning}, our method can scale to these magnitudes with real data.

%

In this paper we tackle this problem by developing an open-source pipeline that vastly reduces
the amount of human annotation time needed to produce labeled RGBD datasets for training image segmentation neural networks. The pipeline
produces ground truth segmentations and ground truth 6DOF poses for multiple objects in scenes with clutter,
occlusions, and varied lighting conditions. The key components of the pipeline are: leveraging dense
RGBD reconstruction to fuse together RGBD images taken from a variety of viewpoints,
labeling with ICP-assisted fitting of object meshes, and automatically rendering labels using projected
object meshes. These techniques allow us to label once per scene, with each
scene containing thousands of images, rather than having to annotate images individually. This  reduces human annotation time by several orders of magnitude over traditional techniques.
We optimize our pipeline to both collect
many views of a scene and to collect many scenes with varied object arrangements.
Our goal is to enable manipulation researchers and practitioners to generate customized datasets, which for example can be used to train any of the available state-of-the-art image segmentation
neural network architectures. Using this method we have
collected over 1,000,000
labeled object instances in multi-object scenes, with only a few days of data collection
and without using any crowd sourcing platforms for human annotation.

Our primary contribution is the pipeline to rapidly generate labeled data, which researchers can use to
build their own datasets, with 
the only hardware requirement being the RGBD sensor itself.  We also have made available our own
dataset, which 
is the largest available RGBD dataset with object-pose labels (352,000 labeled images, 1,000,000+ object
instances).
Additionally, we contribute a number of empirical results concerning the use of large datasets
for practical deep-learning-based pixelwise segmentation of manipulation-relevant scenes in clutter --
specifically, 
we empirically quantify the generalization value of varying aspects of the training data: (i) multi-object vs single
object scenes, (ii) the number of background environments, and (iii) the number
of views per scene. 


%
%


\section{Related Work}

We review three areas of related work.  First, we review pipelines for generating labeled RGBD data.
Second, we review applications of this type of labeled data 
to 6DOF
object pose estimation in the context of robotic manipulation tasks.  Third, we review work related to our
empirical evaluations, concerning questions of scale and generalization for practical learning in
robotics-relevant contexts.

\subsection{Methods for Generating Labeled RGBD Datasets}
Rather than evaluate RGBD datasets based on the specific dataset they provide, we evaluate the
methods used to
generate them, and how well they scale. Firman \cite{firman-cvprw-2016}
provides an extensive overview of over 100 available RGBD datasets. Only a few of the methods used
are capable of generating labels for 6DOF object poses, and none of
these associated datasets also provide per-pixel labeling of objects. 
One of the most related methods to ours is that used to create the T-LESS dataset \cite{hodan2017t},
which contains approximately 49K RGBD images of textureless objects labeled with the 6DOF pose of
each object. 
Compared to our approach, \cite{hodan2017t} requires highly calibrated data collection equipment. They
employ fiducials for camera pose tracking which limits the ability of their method to
 operate in arbitrary environments. 
 Additionally the alignment of the object models to the pointcloud is a completely manual process
 with no algorithmic assistance. 
Similarly, \cite{Wong17} describes a high-precision motion-capture-based approach, 
which
does have the benefit of generating high-fidelity ground-truth pose, but its ability to scale
to large scale data generation is limited by: the confines of the motion capture studio, motion capture
markers on objects interfering with the data collection, and time-intensive setup for each object. 

Although the approach is not capable of generating the 6 DOF poses of objects, a relevant method for per-pixel 
labeling is described in \cite{zeng2016multi}. They employ an automated data collection pipeline
in which
the key idea is to use background subtraction. Two images are taken with the camera at the 
exact same location -- in the first, no object is present, while it is in the second. Background 
subtraction automatically yields a pixelwise segmentation of the object. 
Using this approach they generate 130,000 labeled images for their 39 objects. As a pixelwise labeling method, 
there are a few drawbacks to this 
approach. The first is that in order to 
apply the background subtraction method, they only have a single
object present in each scene. In particular there are no training images with occlusions.  They could in
theory extend their method to support multi-object scenes by adding 
objects to the scene one-by-one, but this presents practical challenges. 
Secondly the approach requires an 
accurately calibrated robot arm to move the camera in a repeatable way. 
A benefit of the method, however, is that
it does enable pixelwise labeling of even deformable objects.



The SceneNN \cite{hua2016scenenn} and ScanNet \cite{dai2017scannet} data
generation pipelines share some features with our method. They both use an RGBD sensor
to produce a dense 3D reconstruction and then perform annotations in 3D. However, since SceneNN and ScanNet are focused on producing datasets for RGDB scene understanding tasks, the type of
annotation that is needed is quite different. In particular their methods provide pixelwise segmenation into generic object classes (floor, wall, couch etc.). Neither SceneNN or ScaneNet
have gometric models for the specific objects in a scene and thus cannot provide 6DOF object poses.
Whereas ScanNet and SceneNN focus on producing datasets for benchmarking scene understanding
algorithms, we provide a pipeline to enable rapid generation labeled data for your particular application and object set.

\subsection{Object-Specific Pose Estimation in Clutter for Robotic Manipulation}

There have been a wide variety of methods to estimate object poses for manipulation.   A challenge is object specificity.  \cite{Wong17}
and \cite{zeng2016multi} are both state of the art pipelines for estimating object poses from RGBD images in clutter -- 
both approaches use RGB
pixelwise segmentation neural networks (trained on their datasets described in the previous section) to crop point clouds which are then
fed into ICP-based algorithms to estimate object poses by registering against prior known meshes.  Another approach is to directly learn pose estimation
\cite{yu2013vision}. The upcoming SIXD Challenge 2017 \cite{SIXD} will provide a comparison of 
state of the art methods for 6DOF pose estimation on a common dataset. The challenge dataset 
contains RGBD images annotated with ground truth 6DOF object poses. This is exactly the type 
of data produced by our pipeline and we aim aim to submit our dataset to the 2018 challenge. 
There is also a trend in manipulation research to bypass object pose estimation and work 
directly with the raw sensor data \cite{levine2016end,gualtieri2016high,mahler2017dex}. 
Making these methods
object-specific in clutter could be aided by using the pipeline presented here to train segmentation networks.

\subsection{Empirical Evaluations of Data Requirements for Image Segmentation Generalization}

While the research community is more familiar with the scale and variety of data needed for images in the style of ImageNet \cite{russakovsky2015imagenet}, the type of visual data that robots have available is much different than ImageNet-style images.  Additionally, higher object specificity may be desired.  In robotics contexts, there has been recent work in trying to identify data requirements for achieving practical performance for deep visual models trained on simulation data \cite{richter2016playing,Johnson-Roberson:2017aa,RosCVPR16,hattori2015learning}, and specifically augmenting small datasets of real data with large datasets of simulation data \cite{richter2016playing,Johnson-Roberson:2017aa,RosCVPR16,hattori2015learning}.  We do not know of prior studies that have performed generalization experiments with the scale of real data used here.

\section{Data Generation Pipeline}

One of the main contributions of this paper is an efficient pipeline for generating labeled RGBD training data.  The steps of the pipeline are described in the following sections: RGBD data collection, dense 3D reconstruction, object mesh generation, human assisted annotation, and rendering of labeled images.

\begin{figure*}[h]
\begin{subfigure}{.33\textwidth}
  \centering
  \includegraphics[width=1.0\linewidth]{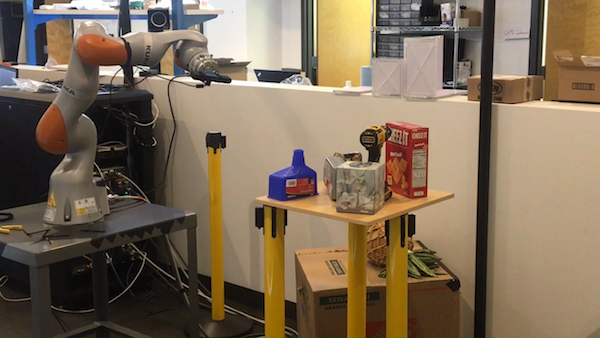}
  \caption{}
  \label{fig:pipieline-sfig1}
\end{subfigure}
\begin{subfigure}{.33\textwidth}
  \centering
  \includegraphics[width=0.8\linewidth]{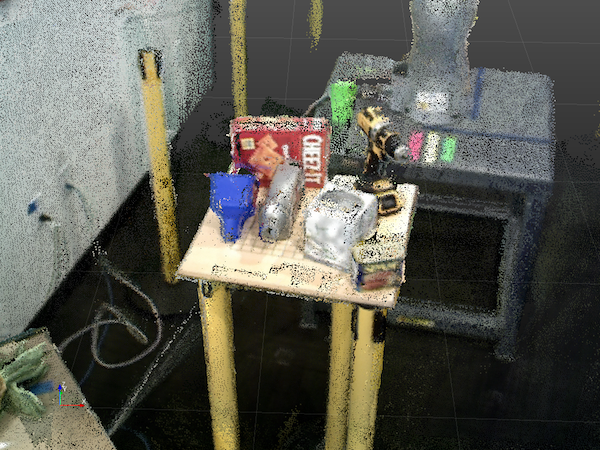}
  \caption{}
  \label{fig:pipieline-sfig2}
\end{subfigure} 
\begin{subfigure}{.33\textwidth}
  \centering
  \includegraphics[width=1.0\linewidth]{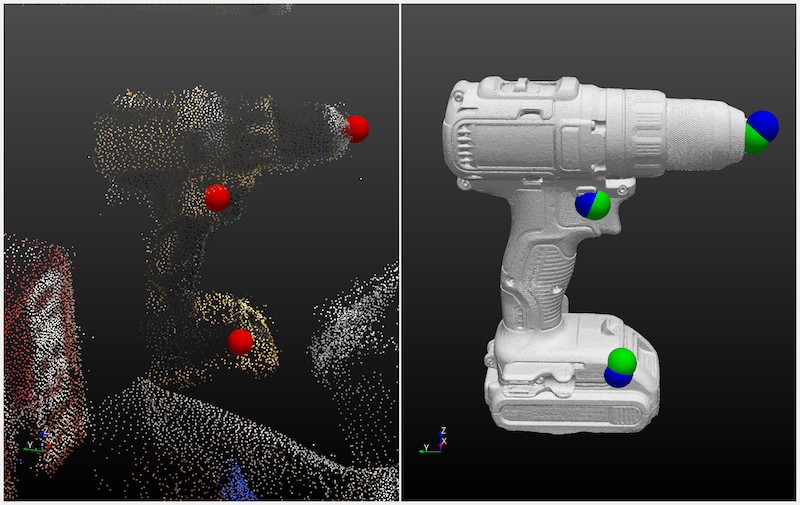}
  \caption{}
  \label{fig:pipieline-sfig3}
\end{subfigure}
\begin{subfigure}{.33\textwidth}
  \centering
  \includegraphics[width=0.9\linewidth]{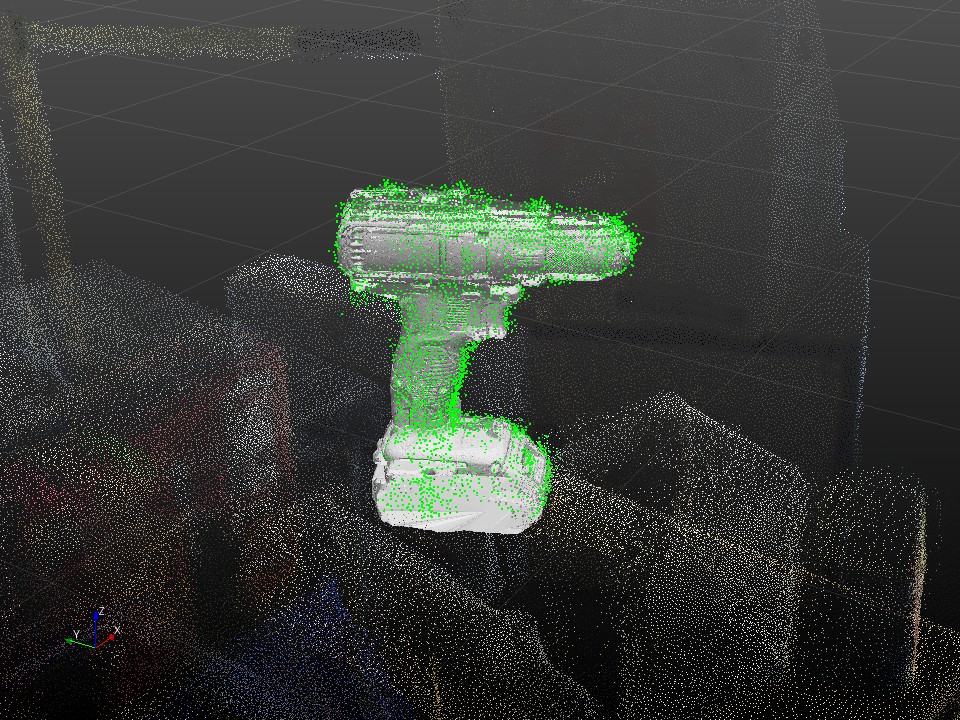}
  \caption{}
  \label{fig:pipieline-sfig4}
\end{subfigure} 
\begin{subfigure}{.33\textwidth}
  \centering
  \includegraphics[width=0.9\linewidth]{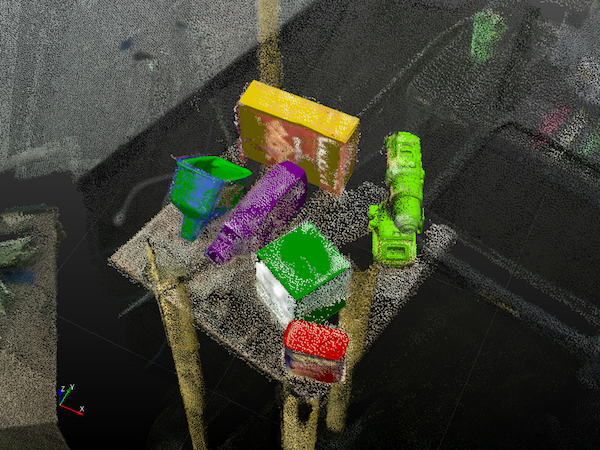}
  \caption{}
  \label{fig:pipieline-sfig5}
\end{subfigure} 
\begin{subfigure}{.33\textwidth}
  \centering
  \includegraphics[width=1.0\linewidth]{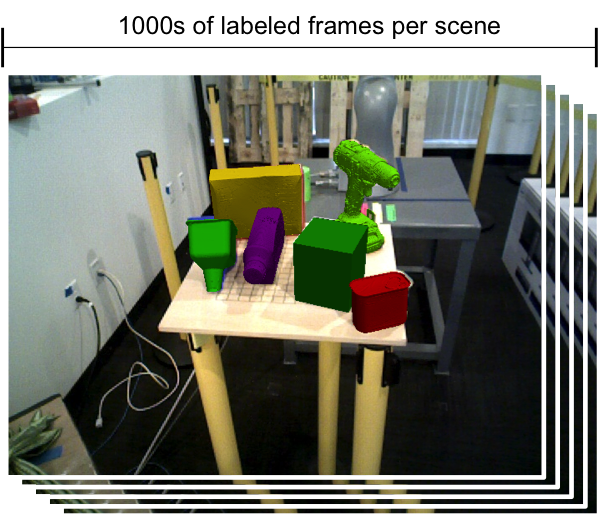}
  \caption{}
  \label{fig:pipieline-sfig6}
\end{subfigure}
\caption{Overview of the data generation pipeline. (a) Xtion RGBD sensor mounted on Kuka IIWA arm for raw data collection. (b) RGBD data processed by ElasticFusion into reconstructed pointcloud. (c) User annotation tool that allows for easy alignment using 3 clicks. User clicks are shown as red and blue spheres. The transform mapping the red spheres to the green spheres is then the user specified guess. (d) Cropped pointcloud coming from user specified pose estimate is shown in green. The mesh model shown in grey is then finely aligned using ICP on the cropped pointcloud and starting from the user provided guess. (e) All the aligned meshes shown in reconstructed pointcloud. (f) The aligned meshes are rendered as masks in the RGB image, producing pixelwise labeled RGBD images for each view.} 
\end{figure*}

\subsection{RGBD Data Collection}
A feature of our approach is that the RGBD sensor can either be mounted on an automated arm, as in Figure (\ref{fig:pipieline-sfig1}), or
the the RGBD sensor can simply be hand-carried.  The benefit of the former option is a reduced human workload, 
while the benefit of the latter option is that no sophisticated equipment (i.e. motion capture, external
markers, heavy robot arm) is required, enabling data collection in a wide variety of environments. 
We captured 112 scenes using the
handheld approach. For the remaining 26 scenes we mounted the sensor on a Kuka IIWA, 
as shown in Figure (\ref{fig:pipieline-sfig1}). The IIWA was programmed to perform a scanning pattern in both
orientation and azimuth.
Note that the arm-automated method does not
require one to know the transform between the robot and the camera; everything is done in camera frame.
Our typical logs averaged 120 seconds in duration with data captured at 30Hz by the Asus Xtion Pro. 

\subsection{Dense 3D Reconstruction}
The next step is to extract a dense 3D reconstruction of the scene, shown in Figure (\ref{fig:pipieline-sfig2}), from the raw RGBD data. 
For this step we used the open source implementation of ElasticFusion \cite{whelan2015elasticfusion} with the default
parameter settings, which runs in realtime on our desktop with an NVIDIA
GTX 1080 GPU. ElasticFusion also provides camera pose tracking relative to the local reconstruction frame, a fact that we take advantage of when rendering labeled images. Reconstruction performance can be affected by the amount of geometric features and RGB texture in the
scene. 
Most natural indoor scenes provide sufficient texture, but large, flat surfaces 
with no RGB or depth texture can occasionally incur failure modes. Our pipeline is designed in a modular 
fashion so that any 3D reconstruction method that provides camera pose tracking can be used in place of ElasticFusion.


\subsection{Object Mesh Generation}




A pre-processing step for the pipeline is to obtain meshes for each object.  Once obtained, meshes speed annotation by enabling alignment of the mesh model rather than manually intensive pixelwise segmentation of the 3D reconstruction.  Using meshes necessitates rigid objects, but imposes no other restrictions on the objects 
themselves. 
We tested several different mesh construction techniques when building our dataset. 
In total there are twelve objects. Four object meshes were generated using 
an Artec Space Spider handheld scanner. One object was scanned using Next Engine turntable scanner. 
For the four objects which are part of the YCB dataset \cite{YCB:2017} we used the provided meshes.  One of our objects, a tissue box, 
was modeled using primitive box geometry. In addition our pipeline provides a volumetric 
meshing method using the VTK implementation of \cite{curless1996volumetric} that operates 
directly on the data already 
produced by ElasticFusion. Finally, there exist several relatively low cost all-in-one solutions \cite{itSeez3D}, \cite{SenseForRealSense}, \cite{Skanect} 
which use RGBD sensors such as the Asus Xtion, Intel RealSense R300 and Occipital Structure
Sensor, to generate object meshes.
The only requirement 
is that the mesh be sufficiently high quality to enable the ICP based alignment (see section 
\ref{Human Assisted Annotation}). RGB textures of meshes are not necessary.  

\subsection{Human Assisted Annotation} \label{Human Assisted Annotation}
One of the key contributions of the paper is in reducing the amount of human annotation time needed to
generate labeled per-pixel and pose data of objects in clutter. 
We evaluated several global registration methods \cite{zhou2016fast,Yang:2016,CGF:CGF12446} to 
try to automatically align our known objects to the 3D
reconstruction but none of them came close to providing satisfactory results. This is due to a variety of
reasons, but a principle one is that many scene points didn't belong to any of the objects.

To circumvent this problem we developed a novel user interface that utilizes human input to assist
traditional registration techniques. The user interface was developed using Director \cite{director}, a
robotics interface and visualization framework. 
Typically the objects of interest are on a table or another
flat surface -- if so, a single click from the user segments out the table.
The user identifies each object in the scene and then selects the corresponding mesh from the mesh library.
They then perform a 3-click-based initialization of the object pose.  Our insight for the alignment stage was that if the user provides a rough initial pose for the object, then 
traditional ICP-based techniques can successfully provide the fine alignment. The
human provides the rough initial alignment by clicking three points on the object in the reconstructed
pointcloud, and then clicking roughly the same three points in the object mesh, see Figure (\ref{fig:pipieline-sfig3}). The transform that best aligns the 3 model points, shown in red, with the three scene points, shown in blue, in a least squares sense is found using the vtkLandmarkTransform function. The resulting transform then specifies an initial alignment of the object
mesh to the scene, and a cropped pointcloud is taken from the points within 1cm of the
roughly aligned model, as shown in green in Figure (\ref{fig:pipieline-sfig4}). Finally, we perform ICP to align this cropped pointcloud to the model, using the rough aligment of the model as the initial seed. In practice this results in very good alignments even for cluttered scenes such as Figure (\ref{fig:pipieline-sfig5}).

The entire human annotation process takes approximately 30 seconds
per object. This is much faster than aligning the full object meshes by hand without using the
3-click technique which can take several minutes per object and results in less accurate object poses.
We also compared our method with human labeling (polygon-drawing) each image, and found intersection over union (IoU) above 80\%, with approximately four orders of magnitude less human effort per image (supplementary figures on our website).

\subsection{Rendering of Labeled Images and Object Poses}
After the human annotation step of Section \ref{Human Assisted Annotation}, the rest of the pipeline is
automated. 
Given the previous steps it is easy to generate
per-pixel object labels by projecting the 3D object poses back into the 2D RGB images. Since our reconstruction method, ElasticFusion, provides camera poses relative to the local reconstruction frame, and we have
already aligned our object models to the reconstructed pointcloud, 
we also have object poses in each camera frame, for each image frame in the log. Given object poses in
camera frame it is easy to get the pixelwise labels by projecting the object meshes into the rendered images. 
An RGB image with projected object meshes is displayed in Figure (\ref{fig:pipieline-sfig6}).

\subsection{Discussion}
As compared to existing methods such as \cite{hodan2017t,DBLP:journals/corr/RennieSBS15,Wong17} our
method requires no sophisticated calibration, works for arbitrary rigid objects in general environments,
and
requires only 30 seconds of human annotation time per object per scene. 
Since the human annotation
is done on the full 3D reconstruction, one labeling effort automatically 
labels thousands of RGBD images of the same scene from different viewpoints.

\section{Results}

We first analyze the effectiveness of the LabelFusion data generation pipeline (Section
\ref{example_data}). We then use data generated from our pipeline to perform
practical empirical experiments to quantify the generalization value
of different aspects of training data (Section \ref{empirical_eval}).

\begin{figure*}[h]
\begin{subfigure}{1.0\textwidth}
  \centering
  \includegraphics[width=1.0\linewidth]{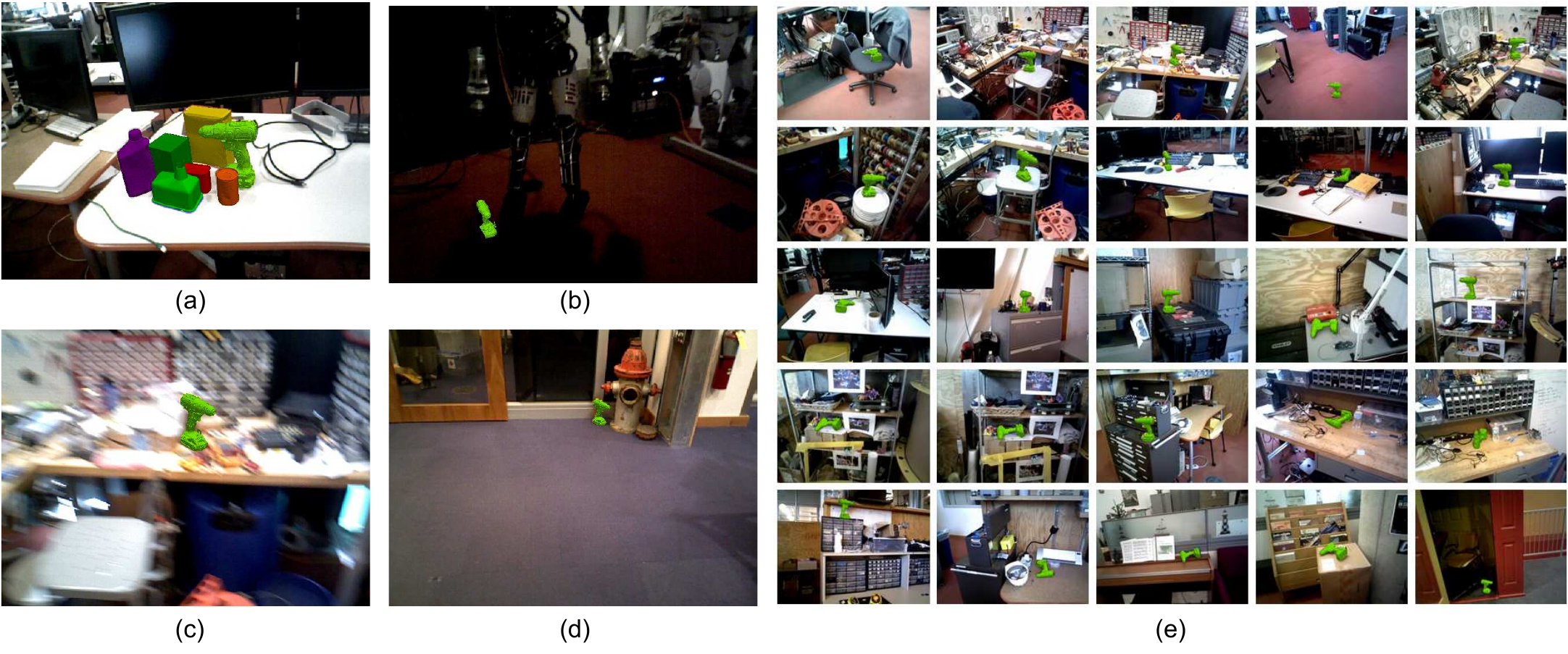}
  \label{fig:drill_city}
\end{subfigure}
\caption{Examples of labeled data generated by our pipeline: (a) heavily cluttered multi-object, (b) low light conditions, (c) motion blur, (d) distance from object, (e) 25 different environments.  All of these scenes were collected by hand-carrying the RGBD sensor.} 
\label{fig:example_data}
\end{figure*}


\subsection{Evaluation of Data Generation Pipeline}\label{example_data}
\begin{table}[]
\centering
\begin{tabular}{|l|l|}
\hline
\# objects                         & 12                                  \\ \hline
\# distinct scenes                 & 105 single/double object\\ & 33 with 6+ objects \\ \hline
\# unique object instances aligned & 339 						             \\ \hline
avg duration of single scene       & 120 seconds, 3600 RGBD frames       \\ \hline
\# labeled RGBD frames             & 352,000                             \\ \hline
\# labeled object instances        & 1,000,000+                          \\ \hline
\end{tabular}
\caption{Dataset Description}
\label{table:dataset_description}
\end{table}

LabelFusion has the capability to rapidly produce large amounts of labeled data, with minimal human
annotation time.  In total we generated over 352,000 labeled RGBD images, of which over 200,000
were generated in approximately one day by two people.  Because many of our images are multi-
object, this amounts to over 1,000,000 labeled object instances. Detailed statistics are provided in Table \ref{table:dataset_description}. The pipeline is open-source and
intended for use. We were able to create training data in a wide variety of scenarios; examples are provided in Figure \ref{fig:example_data}.  
  In particular, we highlight
the wide diversity of environments enabled by hand-carried data collection, the wide variety of lighting
conditions, and the heavy clutter both of backgrounds and of multi-labeled object scenes.



\begin{center}
  \includegraphics[keepaspectratio=true,scale=0.4]{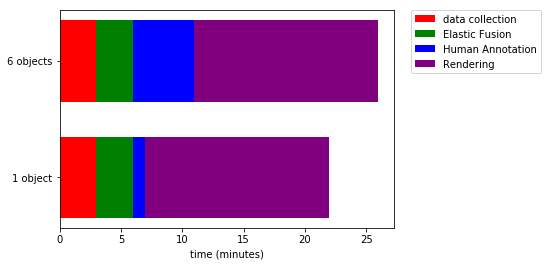}
  \captionof{figure}{Time required for each step of pipeline.} \label{fig:time_analysis}
\end{center}

\begin{center}
  \includegraphics[keepaspectratio=true,scale=0.4]{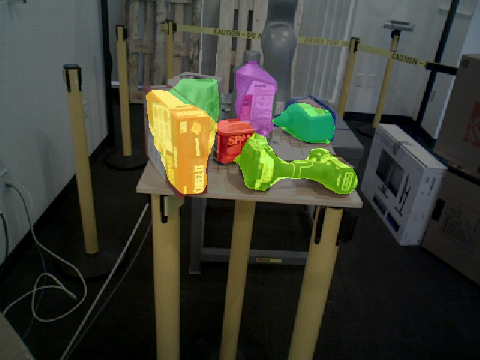}
  \captionof{figure}{Example segmentation performance (alpha-blended with RGB image) of network $(e)$ on a multi-object test scene.}\label{fig:test2}
\end{center}


For scaling to large scale data collection, the time required to generate data is critical. Our pipeline is highly automated and most components run at approximately real-time, as shown in Figure \ref{fig:time_analysis}.  
The amount of human time required is approximately 30 seconds per object per scene, which for a typical single-object scene is less than real-time.  Post-processing runtime is several times greater than real-time, but is easily parallelizable -- in practice, a small cluster of 2-4 modern desktop machines (quad-core Intel i7 and Nvidia GTX 900 series or higher) can be made to post-process the data from a single sensor at real-time rates.  With a reasonable amount of resources (one to two people and a handful of computers), it would be possible to keep up with the real-time rate of the sensor (generating labeled data at 30 Hz). 

\subsection{Empirical Evaluations: How Much Data Is Needed For Practical Object-Specific Segmentation?}\label{empirical_eval}

With the capability to rapidly generate a vast sum of labeled real RGBD data, questions of ``how much data is needed?'' and ``which types of data are most valuable?'' are accessible.  
We explore practical generalization performance while varying three axes of the training data: (i) whether the training set includes multi-object scenes with occlusions or only single-object scenes, (ii) the number of background environments, and (iii) the number of views used per scene.
For each, we train a state-of-the-art ResNet segmentation network \cite{CP2016Deeplab} with different subsets of training data, and evaluate each network's generalization performance on common test sets.  
Further experimental details are provided in our supplementary material; due to space constraints we can only summarize results here. 

\begin{figure*}[h]
\resizebox{\textwidth}{!}{
	\includegraphics[height=4cm]{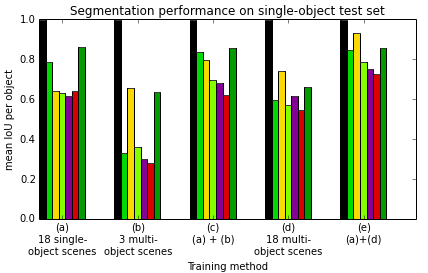}
	\includegraphics[height=4cm]{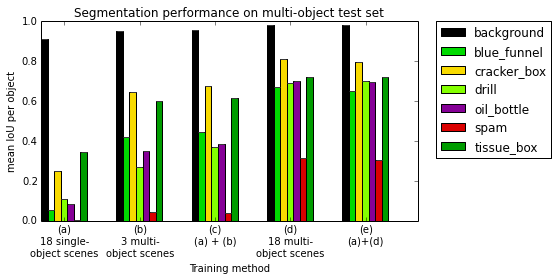}
}
\caption{Comparisons of training on single-object vs. multi-object scenes and testing on single-object (left) and multi-object (right) scenes.} \label{fig:single_multi}
\end{figure*}


First, we investigate whether there is a benefit of using training data with heavily occluded and cluttered multi-object scenes, compared to training with only single-object scenes.  Although they encounter difficulties with heavy occlusions in multi-object scenes, \cite{zeng2016multi} uses purely single-object scenes for training.   We trained five different networks to enable comparison of segmentation performance on novel scenes (different placements of the objects) for a single background environment.  
Results of segmentation performance on novel scenes (measured using the mean IoU, intersection over union, per object) show an advantage given multi-object occluded scenes $(b)$ compared to single-object scenes $(a)$ (Figure \ref{fig:single_multi}, right).  In particular, the average IoU per object increases 190\% given training set $(b)$ instead of $(a)$ in Figure \ref{fig:single_multi}, right, even though $(b)$ has strictly less labeled pixels than $(a)$, due to occlusions.
This implies that the value of the multi-object training data is more valuable per pixel than the single-object training data.  When the same amount of scenes for the single-object scenes are used to train a network with multi-object scenes $(d)$, the increase in IoU performance averaged across objects is 369\%.  Once the network has been trained on 18 multi-object scenes $(d)$, an additional 18 single-object training scenes have no noticeable effect on multi-object generalization $(e)$.  For generalization performance on single-object scenes (Figure \ref{fig:single_multi}, left), this effect is not observed; single-object training scenes are sufficient for IoU performance above 60\%.

\begin{figure*}
  \includegraphics[width=\textwidth]{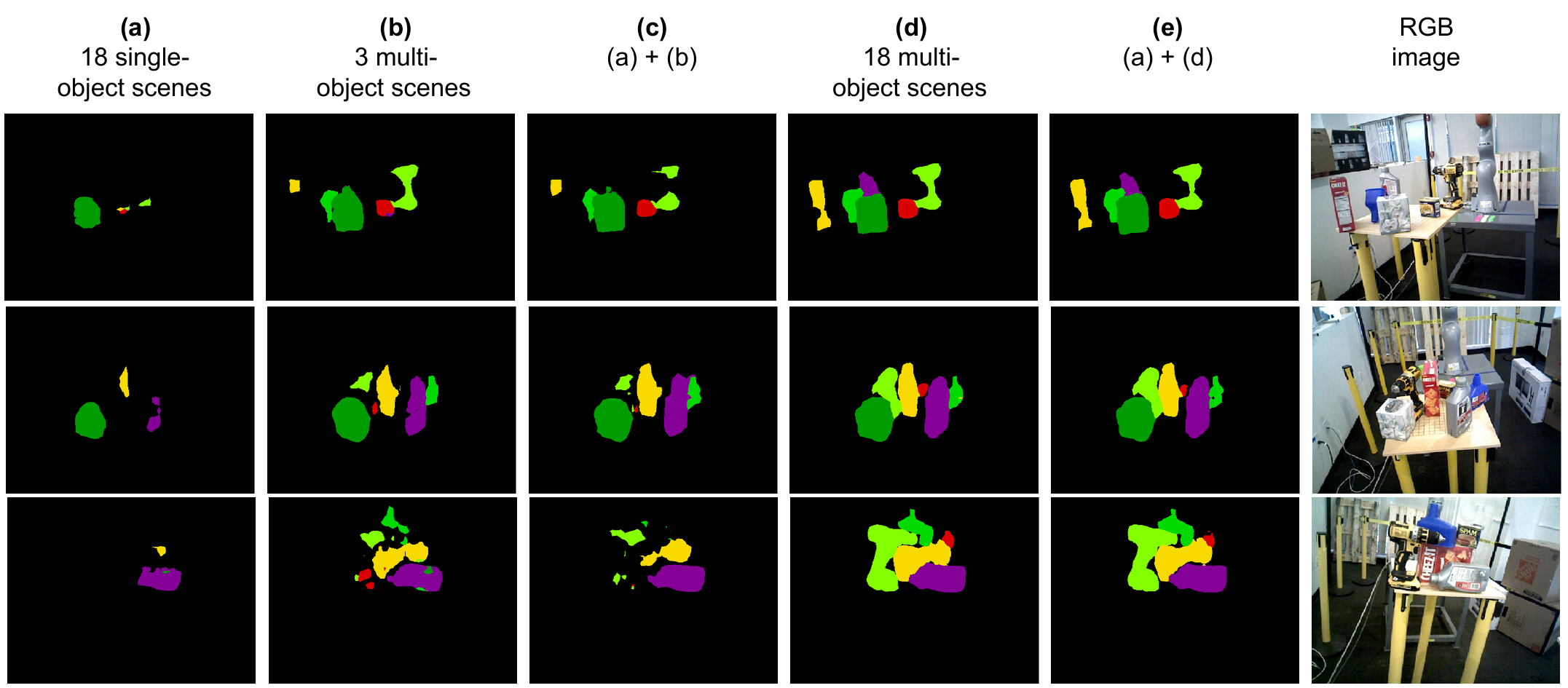}
\caption{Comparison of segmentation performance on novel multi-object test scenes.  Networks are either trained on (a) single object scenes only, (b,d), multi-object test scenes only, or a mixture (c,e).} 
\label{fig:single_multi_learning}

\end{figure*}

\begin{figure*}
\resizebox{\textwidth}{!}{%
	\includegraphics[height=4cm]{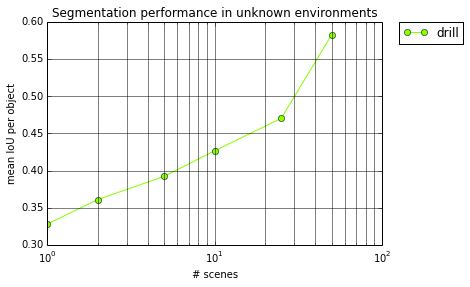}	
	\includegraphics[height=4cm]{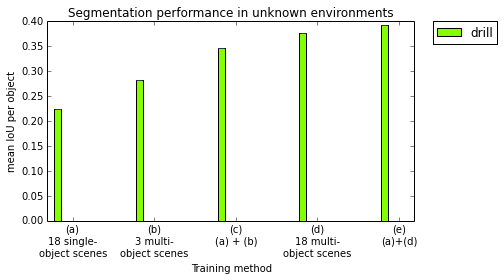}
}
\captionof{figure}{(left) Generalization performance as a function of the number of environments provided at training time, for a set of six networks trained on 50 different scenes or some subset (\{1, 2, 5, 10, 25\}) of those scenes. (right) Performance on the same test set of unknown scenes, but measured for the 5 training configurations for the multi-object, single-environment-only setup described previously.}\label{fig:drill_data}
\end{figure*}

Second, we ask: how does the performance curve grow as more and more training data is added from different background environments?  
To test this, we train different networks respectively on 1, 2, 5, 10, 25, and 50 scenes each labeled with a single drill object. The smaller datasets are subsets of the larger datasets; this directly allows us to measure the value of providing more data.  The test set is comprised of 11 background environments which none of the networks have seen.  We observe a steady increase in segmentation performance that is approximately logarithmic with the number of training scene backgrounds used (Figure \ref{fig:drill_data}, left).  We also took our multi-object networks trained on a single background and tested them on the 11 novel environments with the drill.  We observe an advantage of the multi-object training data with occlusions over the single-object training data in generalizing to novel background environments (Figure \ref{fig:drill_data}, right).

\begin{figure*}
\begin{subfigure}{1.0\textwidth}
  \centering
  \includegraphics[width=1.0\linewidth]{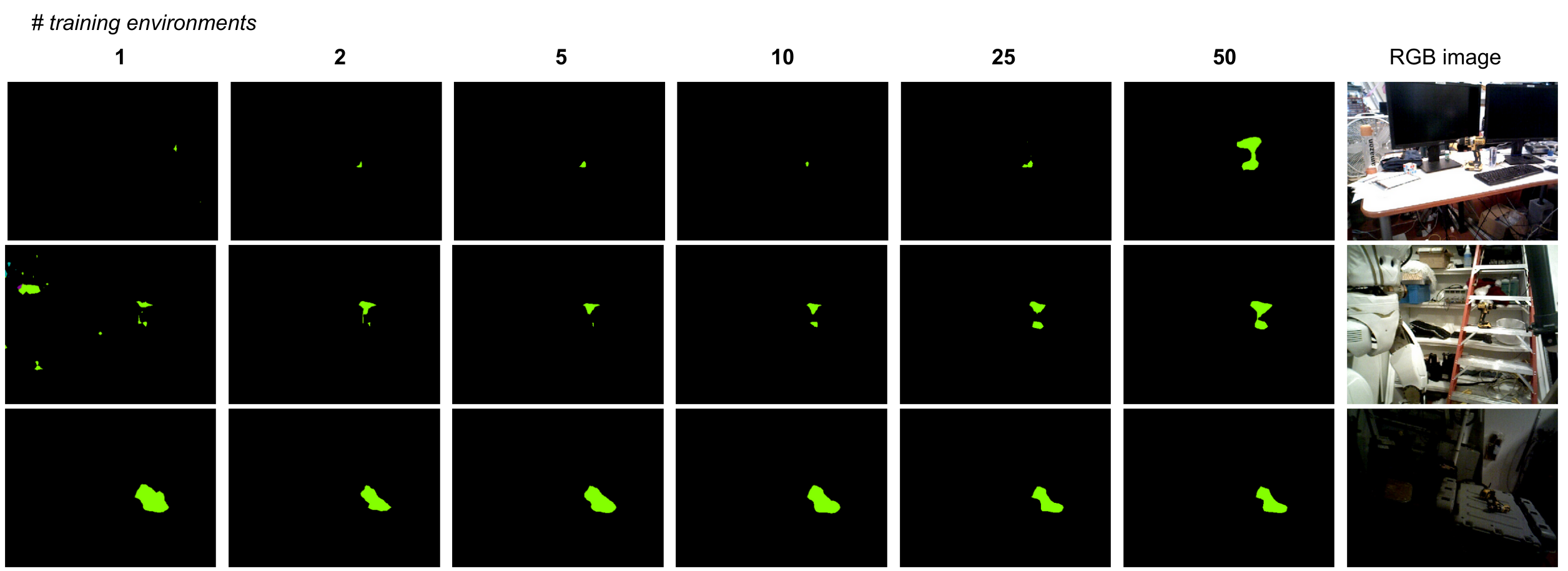}
\end{subfigure}
\caption{Comparison of segmentation performance on novel background environments.  Networks were trained on \{1, 2, 5, 10, 25, 50\} background environments.} 
\label{fig:drill_learning}
\end{figure*}

\begin{figure*}
\begin{subfigure}{1.0\textwidth}
  \centering
  \includegraphics[width=1.0\linewidth]{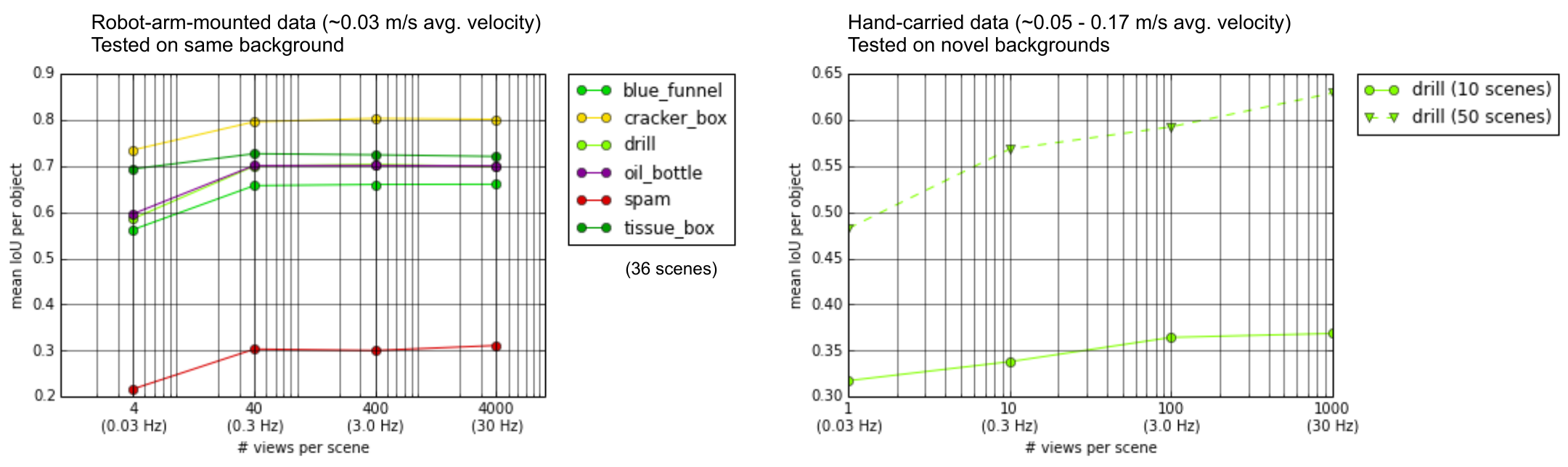}
\end{subfigure}
\caption{Pixelwise segmentation performance as a function of the number of views per scene, reduced by downsampling the native 30 Hz sensor to \{0.03, 0.3, 3.0.\} Hz.} 
\label{fig:hz_learning}
\end{figure*}

Third, we investigate whether 30 Hz data is necessary, 
or whether significantly less data suffices (Figure \ref{fig:hz_learning}).  
We perform experiments with downsampling the effective sensor rate
both for robot-arm-mounted multi-object single-background training set $(e)$, and the hand-carried many-environments dataset with either 10 or 50 scenes.
For each, we train four different networks, where one has all data available and the others have
downsampled data at respectively 0.03, 0.3, and 3 Hz.
We observe a monotonic increase in segmentation performance as the effective sensor rate is
increased, but with heavily diminished returns after 0.3 Hz for the slower robot-arm-mounted data ($\sim$0.03 m/s camera motion velocity). The hand-carried data ($\sim$0.05 - 0.17 m/s) shows more gains with higher rates.

\section{Conclusion}
This paper introduces LabelFusion, our pipeline for efficiently
generating RGBD data annotated with per-pixel labels and ground truth object poses. 
Specifically only a few minutes of human time are required for 
labeling a scene containing thousands of RGBD images. LabelFusion is open source and available for 
community use, and we also supply an example dataset
generated by our pipeline \cite{labelfusion}.

The capability to produce a large, labeled dataset enabled us
to answer several questions related to the type and quantity of training data needed for 
practical deep learning segmentation networks in a robotic manipulation 
context. Specifically we found that networks trained on multi-object scenes performed 
significantly better than those trained on single object scenes, both on 
novel multi-object scenes with 
the same background, and on 
single-object scenes with new backgrounds. Increasing the variety of backgrounds in the 
training data for single-object scenes also improved generalization performance for new backgrounds, with approximately 50 different backgrounds breaking into above-50\% IoU on entirely novel scenes.  
Our recommendation is to focus on multi-object data collection in a variety of backgrounds for the
most gains in generalization performance.

We hope that our pipeline lowers the barrier to entry for using deep learning 
approaches for perception in support of robotic manipulation tasks by reducing 
the amount of human time needed to generate vast quantities of labeled data for 
\emph{your} specific environment and set of objects. It is also our hope that our analysis of 
segmentation network performance provides guidance on the type and quantity of 
data that needs to be collected to achieve desired levels of generalization performance.

%



\section*{ACKNOWLEDGEMENT}
The authors thank Matthew O'Kelly for his guidance
with segmentation networks and manuscript feedback.
We also thank Allison Fastman and Sammy Creasey 
of Toyota Research Institute for their help with hardware, including
object scanning and robot arm automation.  David Johnson of Draper Laboratory
and Shuran Song of Princeton University provided valuable input
on training.  We are grateful we were able to use the robot arm testing facility
from Toyota Research Insitute.
This work was supported by the Air Force/Lincoln Laboratory award no. 7000374874, by the Defense Advanced Research Projects
Agency via Air Force Research Laboratory award FA8750-12-1-0321, and
by NSF Contract IIS-1427050.  The views expressed are not endorsed by the sponsors.


\bibliographystyle{IEEEtran}
\bibliography{corl-bib.bib}  

\begin{thebibliography}{10}
\providecommand{\url}[1]{#1}
\csname url@samestyle\endcsname
\providecommand{\newblock}{\relax}
\providecommand{\bibinfo}[2]{#2}
\providecommand{\BIBentrySTDinterwordspacing}{\spaceskip=0pt\relax}
\providecommand{\BIBentryALTinterwordstretchfactor}{4}
\providecommand{\BIBentryALTinterwordspacing}{\spaceskip=\fontdimen2\font plus
\BIBentryALTinterwordstretchfactor\fontdimen3\font minus
  \fontdimen4\font\relax}
\providecommand{\BIBforeignlanguage}[2]{{%
\expandafter\ifx\csname l@#1\endcsname\relax
\typeout{** WARNING: IEEEtran.bst: No hyphenation pattern has been}%
\typeout{** loaded for the language `#1'. Using the pattern for}%
\typeout{** the default language instead.}%
\else
\language=\csname l@#1\endcsname
\fi
#2}}
\providecommand{\BIBdecl}{\relax}
\BIBdecl

\bibitem{Wong17}
J.~M. {Wong}, V.~{Kee}, T.~{Le}, S.~{Wagner}, G.-L. {Mariottini},
  A.~{Schneider}, L.~{Hamilton}, R.~{Chipalkatty}, M.~{Hebert}, D.~M.~S.
  {Johnson}, J.~{Wu}, B.~{Zhou}, and A.~{Torralba}, ``{SegICP: Integrated Deep
  Semantic Segmentation and Pose Estimation},'' \emph{ArXiv e-prints}, Mar.
  2017.

\bibitem{zeng2016multi}
A.~Zeng, K.-T. Yu, S.~Song, D.~Suo, E.~Walker~Jr, A.~Rodriguez, and J.~Xiao,
  ``Multi-view self-supervised deep learning for 6d pose estimation in the
  amazon picking challenge,'' \emph{arXiv preprint arXiv:1609.09475}, 2016.

\bibitem{richter2016playing}
S.~R. Richter, V.~Vineet, S.~Roth, and V.~Koltun, ``Playing for data: Ground
  truth from computer games,'' in \emph{European Conference on Computer
  Vision}.\hskip 1em plus 0.5em minus 0.4em\relax Springer, 2016, pp. 102--118.

\bibitem{Johnson-Roberson:2017aa}
M.~Johnson-Roberson, C.~Barto, R.~Mehta, S.~N. Sridhar, K.~Rosaen, and
  R.~Vasudevan, ``Driving in the matrix: Can virtual worlds replace
  human-generated annotations for real world tasks?'' in \emph{{IEEE}
  International Conference on Robotics and Automation}, 2017, pp. 1--8.

\bibitem{RosCVPR16}
G.~Ros, L.~Sellart, J.~Materzynska, D.~Vazquez, and A.~Lopez, ``{The SYNTHIA
  Dataset}: A large collection of synthetic images for semantic segmentation of
  urban scenes,'' 2016.

\bibitem{hattori2015learning}
H.~Hattori, V.~Naresh~Boddeti, K.~M. Kitani, and T.~Kanade, ``Learning
  scene-specific pedestrian detectors without real data,'' in \emph{Proceedings
  of the IEEE Conference on Computer Vision and Pattern Recognition}, 2015, pp.
  3819--3827.

\bibitem{firman-cvprw-2016}
M.~Firman, ``{RGBD Datasets: Past, Present and Future},'' in \emph{CVPR
  Workshop on Large Scale 3D Data: Acquisition, Modelling and Analysis}, 2016.

\bibitem{hodan2017t}
T.~Hodan, P.~Haluza, S.~Obdrzalek, J.~Matas, M.~Lourakis, and X.~Zabulis,
  ``T-less: An rgb-d dataset for 6d pose estimation of texture-less objects,''
  \emph{arXiv preprint arXiv:1701.05498}, 2017.

\bibitem{hua2016scenenn}
B.-S. Hua, Q.-H. Pham, D.~T. Nguyen, M.-K. Tran, L.-F. Yu, and S.-K. Yeung,
  ``Scenenn: A scene meshes dataset with annotations,'' in \emph{3D Vision
  (3DV), 2016 Fourth International Conference on}.\hskip 1em plus 0.5em minus
  0.4em\relax IEEE, 2016, pp. 92--101.

\bibitem{dai2017scannet}
A.~Dai, A.~X. Chang, M.~Savva, M.~Halber, T.~Funkhouser, and M.~Nie{\ss}ner,
  ``Scannet: Richly-annotated 3d reconstructions of indoor scenes,''
  \emph{arXiv preprint arXiv:1702.04405}, 2017.

\bibitem{yu2013vision}
J.~Yu, K.~Weng, G.~Liang, and G.~Xie, ``A vision-based robotic grasping system
  using deep learning for 3d object recognition and pose estimation,'' in
  \emph{Robotics and Biomimetics (ROBIO), 2013 IEEE International Conference
  on}.\hskip 1em plus 0.5em minus 0.4em\relax IEEE, 2013, pp. 1175--1180.

\bibitem{SIXD}
\BIBentryALTinterwordspacing
``{SIXD Challenge}.'' [Online]. Available:
  \url{http://cmp.felk.cvut.cz/sixd/challenge_2017/}
\BIBentrySTDinterwordspacing

\bibitem{levine2016end}
S.~Levine, C.~Finn, T.~Darrell, and P.~Abbeel, ``End-to-end training of deep
  visuomotor policies,'' \emph{Journal of Machine Learning Research}, vol.~17,
  no.~39, pp. 1--40, 2016.

\bibitem{gualtieri2016high}
M.~Gualtieri, A.~ten Pas, K.~Saenko, and R.~Platt, ``High precision grasp pose
  detection in dense clutter,'' in \emph{Intelligent Robots and Systems (IROS),
  2016 IEEE/RSJ International Conference on}.\hskip 1em plus 0.5em minus
  0.4em\relax IEEE, 2016, pp. 598--605.

\bibitem{mahler2017dex}
J.~Mahler, J.~Liang, S.~Niyaz, M.~Laskey, R.~Doan, X.~Liu, J.~A. Ojea, and
  K.~Goldberg, ``Dex-net 2.0: Deep learning to plan robust grasps with
  synthetic point clouds and analytic grasp metrics,'' \emph{arXiv preprint
  arXiv:1703.09312}, 2017.

\bibitem{russakovsky2015imagenet}
O.~Russakovsky, J.~Deng, H.~Su, J.~Krause, S.~Satheesh, S.~Ma, Z.~Huang,
  A.~Karpathy, A.~Khosla, M.~Bernstein \emph{et~al.}, ``Imagenet large scale
  visual recognition challenge,'' \emph{International Journal of Computer
  Vision}, vol. 115, no.~3, pp. 211--252, 2015.

\bibitem{whelan2015elasticfusion}
T.~Whelan, S.~Leutenegger, R.~Salas-Moreno, B.~Glocker, and A.~Davison,
  ``Elasticfusion: Dense slam without a pose graph.''\hskip 1em plus 0.5em
  minus 0.4em\relax Robotics: Science and Systems.

\bibitem{YCB:2017}
\BIBentryALTinterwordspacing
B.~Calli, A.~Singh, J.~Bruce, A.~Walsman, K.~Konolige, S.~Srinivasa, P.~Abbeel,
  and A.~M. Dollar, ``Yale-cmu-berkeley dataset for robotic manipulation
  research,'' \emph{The International Journal of Robotics Research}, vol.~36,
  no.~3, pp. 261--268, 2017. [Online]. Available:
  \url{https://doi.org/10.1177/0278364917700714}
\BIBentrySTDinterwordspacing

\bibitem{curless1996volumetric}
B.~Curless and M.~Levoy, ``A volumetric method for building complex models from
  range images,'' in \emph{Proceedings of the 23rd annual conference on
  Computer graphics and interactive techniques}.\hskip 1em plus 0.5em minus
  0.4em\relax ACM, 1996, pp. 303--312.

\bibitem{itSeez3D}
\BIBentryALTinterwordspacing
``{itSeez3D}.'' [Online]. Available: \url{https://itseez3d.com}
\BIBentrySTDinterwordspacing

\bibitem{SenseForRealSense}
\BIBentryALTinterwordspacing
``{Sense for RealSense}.'' [Online]. Available:
  \url{https://www.3dsystems.com/shop/realsense/sense}
\BIBentrySTDinterwordspacing

\bibitem{Skanect}
\BIBentryALTinterwordspacing
``{Skanect}.'' [Online]. Available: \url{http://skanect.occipital.com/}
\BIBentrySTDinterwordspacing

\bibitem{zhou2016fast}
Q.-Y. Zhou, J.~Park, and V.~Koltun, ``Fast global registration,'' in
  \emph{European Conference on Computer Vision}.\hskip 1em plus 0.5em minus
  0.4em\relax Springer, 2016, pp. 766--782.

\bibitem{Yang:2016}
\BIBentryALTinterwordspacing
J.~Yang, H.~Li, D.~Campbell, and Y.~Jia, ``Go-icp: A globally optimal solution
  to 3d icp point-set registration,'' \emph{IEEE Trans. Pattern Anal. Mach.
  Intell.}, vol.~38, no.~11, pp. 2241--2254, Nov. 2016. [Online]. Available:
  \url{https://doi.org/10.1109/TPAMI.2015.2513405}
\BIBentrySTDinterwordspacing

\bibitem{CGF:CGF12446}
\BIBentryALTinterwordspacing
N.~Mellado, D.~Aiger, and N.~J. Mitra, ``Super 4pcs fast global pointcloud
  registration via smart indexing,'' \emph{Computer Graphics Forum}, vol.~33,
  no.~5, pp. 205--215, 2014. [Online]. Available:
  \url{http://dx.doi.org/10.1111/cgf.12446}
\BIBentrySTDinterwordspacing

\bibitem{director}
\BIBentryALTinterwordspacing
P.~Marion, ``Director: A robotics interface and visualization framework,''
  2015. [Online]. Available: \url{http://github.com/RobotLocomotion/director}
\BIBentrySTDinterwordspacing

\bibitem{DBLP:journals/corr/RennieSBS15}
\BIBentryALTinterwordspacing
C.~Rennie, R.~Shome, K.~E. Bekris, and A.~F.~D. Souza, ``A dataset for improved
  rgbd-based object detection and pose estimation for warehouse
  pick-and-place,'' \emph{CoRR}, vol. abs/1509.01277, 2015. [Online].
  Available: \url{http://arxiv.org/abs/1509.01277}
\BIBentrySTDinterwordspacing

\bibitem{CP2016Deeplab}
L.-C. Chen, G.~Papandreou, I.~Kokkinos, K.~Murphy, and A.~L. Yuille, ``Deeplab:
  Semantic image segmentation with deep convolutional nets, atrous convolution,
  and fully connected crfs,'' \emph{arXiv:1606.00915}, 2016.

\bibitem{labelfusion}
``{LabelFusion},'' \url{http://labelfusion.csail.mit.edu}.

\end{thebibliography}

\end{document}